\documentclass[11pt,a4paper]{article}
\usepackage[hyperref]{acl2019}
\usepackage{times}
\usepackage{latexsym}

\def\aclpaperid{} 

\usepackage{multirow}
\usepackage{url}
\usepackage{xspace}
\usepackage{tabularx}
\usepackage{microtype}
\usepackage{graphicx}
\usepackage{paralist}
\usepackage{subfig}
\usepackage{booktabs}
\usepackage{amsmath}
\usepackage{amssymb}
\usepackage{xstring}
\usepackage{scalefnt}
\usepackage{bm}
\usepackage{paralist}
\usepackage{tablefootnote}
\usepackage{latexsym}
\usepackage{calc}
\usepackage{booktabs}
\usepackage{scrextend}
\usepackage[utf8]{inputenc}
\usepackage{enumitem}

\usepackage{hyperref}

\usepackage[all]{nowidow}

\newcommand\footnoteref[1]{\protected@xdef\@thefnmark{\ref{#1}}\@footnotemark}

\newcommand{\posbox}[1]{{\setlength{\fboxsep}{1pt}\colorbox{lightblue}{#1}}}
\newcommand{\negbox}[1]{{\setlength{\fboxsep}{1pt}\colorbox{lightred}{#1}}}

\definecolor{lightgreen}{RGB}{200,255,200}
\definecolor{lightblue}{RGB}{200,200,255}
\definecolor{lightred}{RGB}{255,200,200}

\newcommand{\rt}[1]{\rotatebox{90}{#1}}

\newcommand{\ie}{\textit{i.\,e.}\xspace}
\newcommand{\eg}{\textit{e.\,g.}\xspace}

\aclfinalcopy 

\title{Sentiment analysis is not solved!\\ Assessing and probing sentiment classification}

\author{\textbf{Jeremy Barnes, Lilja Øvrelid, Erik Velldal}\\[5pt]
University of Oslo\\
{\tt \{jeremycb,liljao,erikve\}@ifi.uio.no}
}

\date{}
\begin{document}
\maketitle
\begin{abstract}

Neural methods for SA have led to quantitative improvements over previous approaches, but these advances are not always accompanied with a thorough analysis of the qualitative differences. Therefore, it is not clear what outstanding conceptual challenges for sentiment analysis remain. In this work, we attempt to discover what challenges still prove a problem for sentiment classifiers for English and to provide a challenging dataset. We collect the subset of sentences that an (oracle) ensemble of state-of-the-art sentiment classifiers misclassify and then annotate them for 18 linguistic and paralinguistic phenomena, such as negation, sarcasm, modality, etc.\footnote{The dataset is available at \url{https://github.com/ltgoslo/assessing_and_probing_sentiment}.} Finally, we provide a case study that demonstrates the usefulness of the dataset to probe the performance of a given sentiment classifier with respect to linguistic phenomena.

\end{abstract}

\section{Introduction}

Over the last 15 years, approaches to sentiment analysis which concentrated on creating and curating sentiment lexicons \cite{Turney2002,Liu2005} or used n-grams for classification \cite{Pang2002} have been replaced by models that are able to exploit compositionality \cite{Socher2013b,Irsoy2014a} or implicitly learn relations between tokens \cite{Peters2018,Howard2018,Devlin2018}. These neural models push the state of the art to over 90\% accuracy on binary sentence-level sentiment analysis.

Although these methods show a quantitative improvement over previous approaches, they are not often accompanied with a thorough analysis of the qualitative differences. This has led to the current situation, where we are aware of quantitative, but not qualitative differences between state-of-the-art sentiment classifiers. It also means that we are not
aware of the outstanding conceptual challenges that we still face in sentiment analysis.

In this work, we attempt to discover what conceptual challenges still prove a problem for all state-of-the-art sentiment methods for English. To do so, we train and test three state-of-the-art machine learning classifiers (BERT, ELMo, and a BiLSTM) as well as a bag-of-words classifier on six sentence-level sentiment datasets available for English. We then collect the subset of sentences that all models misclassify and annotate them for 18 linguistic and paralinguistic phenomena, such as negation, sarcasm, modality or world knowledge. We present this new data as a challenging dataset for future research in sentiment analysis, which enables probing the problems that sentiment classifiers still face in more depth.

Specifically, the contributions of this work are:

\begin{itemize}
  \setlength{\itemsep}{5pt}
  \setlength{\parskip}{0pt}
  \setlength{\parsep}{0pt}
\item the creation of a challenging sentiment dataset from previously available data,
\item the annotation of errors in this dataset for 18 linguistic and paralinguistic phenomena,
\item a thorough analysis of the dataset,
\item and finally presenting a practical use-case demonstrating how the dataset can be
  used to probe the particular types of errors made by a new model. 
\end{itemize} 

The rest of the paper is organized into related work (Section \ref{relatedwork}), a description of the experimental setup (Section \ref{expsetup}), a brief description of the dataset (Section \ref{challengedataset}), an in-depth analysis (Section \ref{datasetanalysis}), a case-study that demonstrates the usefulness of the dataset (Section \ref{casestudy}), and finally a conclusion (Section \ref{conclusion}).

\section{Related Work}
\label{relatedwork}

Neural networks are now ubiquitous in NLP tasks, often giving state-of-the-art results. However, they are known for being ``black boxes'' which are not easily interpretable. Recent interest in interpreting these methods has led to new lines of research which attempt to discover what linguistic phenomena neural networks are able to learn \cite{Linzen2016,Gulordava2018,Conneau2018}, how robust neural networks are to perturbations in input data \cite{Ribeiro2018,Ebrahimi2018,Schluter2018}, and what biases they propagate \cite{Park2018,Zhao2018,Kiritchenko2018}. 

Specifically within the task of sentiment analysis, certain linguistic phenomena are known to be challenging. Negation is one of the aspects of language that most clearly affects expressions of sentiment and that has been studied widely within sentiment analysis (see \newcite{Wiegand2010} for an early survey). The difficulties of resolving negation for sentiment analysis include determining negation scope \cite{Hogenboom2011,Lapponi2012,Reitan2015}, and semantic composition \cite{Wilson2005,Choi2008,Kiritchenko2016}.

Verbal polarity shifters have also been studied. \newcite{Schulder2018} annotate verbal shifters at the sense-level. They conclude that, although individual negation words are more frequent in the Amazon Product Review Data corpus, the overall frequency of negation words and shifters is likely similar. This suggests that there is a Zipfian tail of shifters which are not often handled within sentiment analysis.

Furthermore, the linguistic phenomenon of modality has also been shown to be problematic. Both \newcite{Narayanan2009} and \newcite{Liu2014} explore the effect of modality on sentiment classification and find that explicitly modeling certain modalities improves classification results. They advocate for a divide-and-conquer approach, which would address the various realizations of modality individually. \newcite{Benamara2012} perform linguistic experiments using native speakers concerning the effects of both negation and modality on opinions, and similarly find that the type of negation and modality determines the final interpretation of polarity.

The sentiment models inspected in these analyses, however, were lexicon- and word- and n-gram-based models. It is not clear that neural networks have the same weaknesses, as they have been shown to deal with compositionality and long-distance dependencies to some degree \cite{Socher2013b,Linzen2016}. Additionally, authors did not attempt to discover from the data what phenomena were present that could affect sentiment. In the current paper we aim to provide a systematic analysis of error types found across a range of datasets, domains and classifiers.

\section{Experimental Setup}
\label{expsetup}

In these experiments, we test three state-of-the-art models for sentence-level sentiment classification. We choose to focus on sentence-level classification for three reasons: 1) sentence-level classification is a popular and useful task, 2) there is a large amount of high-quality annotated data available, and 3) annotation of linguistic phenomena is easier at sentence-level than document-level. It is also likely that most phenomena that occur at sentence-level, \eg, negation, comparative sentiment, or modality, will transfer to other sentiment tasks.

\subsection{Datasets}

In order to discover a subset of sentences that all state-of-the-art models are unable to correctly predict, we collect six English-language datasets previously annotated for sentence-level sentiment from five domains (news wire, hotel reviews, movie reviews, twitter, and micro-blogs). Table \ref{datasetstats} shows the statistics for each of the datasets.

\begin{table}

\newcommand{\sep}{\cmidrule(lr){2-2}\cmidrule(lr){3-3}\cmidrule(lr){4-4}\cmidrule(lr){5-5}\cmidrule(lr){6-6}\cmidrule(lr){7-7}}

\centering\small
\setlength{\tabcolsep}{5pt}

\begin{tabular}{lrrrrrr}
\toprule
& MPQA & OP. & Sem. & SST & Ta. & Th. \\
\sep
$++$ &  $-$  & 379  &  $-$  & 1,852  &  $-$  & $-$ \\
$+$ &  193   & 879  &  3,499 & 3,111  & 923   & 2,727 \\
0 &    527   &  $-$ &  4,478 & 2,242  & 1,419  & 1,779 \\
$-$ &  413   & 399  &  1,310 & 3,140  & 1,320  & 1,828 \\
$--$ &  $-$  & 74   &  $-$  & 1,510  & $-$   & $-$ \\
\sep
total   & 1,133 & 1,731 & 9,287 & 11,855 & 3,662 & 6,334 \\
\bottomrule
\end{tabular}
\caption{Statistics for the sentence-level annotations in each dataset.}
\label{datasetstats}
\end{table}

\paragraph{MPQA}

The Multi-perspective Question Answer (MPQA) Opinion Corpus \cite{Wilson2005} provides contextual polarity annotations for English news documents from world press. The annotations are private state frames, which include annotations for text anchor, source, target, and attitude type, among others. We extract sentiment labeled sentences by taking only those sentences that have sentiment annotations. Additionally, we remove sentences that contain both positive and negative sentiment. This leaves a three-class (positive, neutral, negative) sentence-level dataset.

\paragraph{OpeNER}

The Open Polarity Enhanced Named Entity Recognition (OpeNER) sentiment datasets \cite{Agerri2013} contain hotel reviews annotated for 4-class (strong positive, positive, negative, strong negative) sentiment classification. We take the English dataset, where self-attention networks give state-of-the-art results \cite{Ambartsoumian2018}.

\paragraph{SemEval}

The SemEval 2013 tweet classification dataset \cite{Nakov2013} contains tweets collected and annotated for three-class (positive, neutral, negative) sentiment. The state-of-the-art model is a Convolutional Network \cite{Severyn2015}.

\paragraph{Stanford Sentiment Treebank}

The Stanford Sentiment Treebank \citep{Socher2013b} contains 11,855 English sentences from movie reviews which have been annotated at each node of a constituency parse tree. Contextualized word representations combined with a bi-attentive sentiment network currently give state-of-the-art results \cite{Peters2018}.

\paragraph{Täckström dataset}

The Täckström dataset \cite{Tackstrom2011} contains product reviews which have been annotated at both document- and sentence-level for three-class sentiment, although the sentence-level annotations also have a ``not relevant'' label. We keep the sentence-level annotations, which gives 3,662 sentences annotated for three-class sentiment. 

\paragraph{Thelwall dataset}

The Thelwall dataset derives from datasets provided with SentiStrength\footnote{The data are available at \url{http://sentistrength.wlv.ac.uk/}} \cite{Thelwall2010}. It contains microblogs annotated for both positive and negative sentiment on a scale from 1 to 5. We map these to single sentiment labels such that sentences which are clearly positive (pos $>=$ 3 and neg $<$ 3) are given the positive label,
clearly negative sentences (pos $<$ 3 and neg $>=$ 3) the negative label, and clearly neutral sentences ( 3 $<$ pos $>$ 2 and 3 $<$ neg $>$ 2) the neutral. We discard all other sentences, which finally leaves 6,334 annotated sentences.

\subsection{Models}

In order to gain an idea of what errors most models suffer from, we test three state-of-the-art models on the datasets. Additionally, we use a bag-of-words model as it is a strong baseline for text classification. For the \textsc{Single} setup, we train all models on the training and development data for each dataset and test on the corresponding test set, therefore avoiding domain problems.

\paragraph{BERT} The BERT model \cite{Devlin2018} is a bidirectional transformer that is pretrained on two tasks: 1) a cloze-like language modeling task and 2) a binary next-sentence prediction task. It is pre-trained on 330 million words from the BooksCorpus \cite{Zhu_2015_ICCV} and English Wikipedia. We fine-tune the available pretrained model\footnote{\url{https://github.com/google-research/bert}} on each sentiment dataset.

\paragraph{ELMo} We use the bi-attentive classification network\footnote{\url{https://s3-us-west-2.amazonaws.com/allennlp/models/sst-5-elmo-biattentive-classification-network-2018.09.04.tar.gz}} from \newcite{Peters2018}. The network uses both word embeddings, as well as creating character-based embeddings from a character-level CNN-BiLSTM network. The word representations are first passed through a feedforward layer, and then through a sequence-to-sequence network with biattention. This new representation of the text is combined with the original representation and passed through another sequence-to-sequence network. Finally, a max, min, mean and self-attention pool representation is created from this last sequence. For classification, these features are sent to a maxout layer.

\paragraph{BiLSTM} Bidirectional long short-term memory (BiLSTM) networks have shown to be strong baselines for sentiment tasks \cite{Tai2015a,Barnes2017}. We implement a single-layered BiLSTM which takes pretrained skipgram embeddings as input, creates a sentence representation by concatenating the final hidden layer of both left and right LSTMs, and then passes this representation to a softmax layer for classification. Additionally, dropout serves as a regularizer.

\paragraph{Bag-of-Words classifier} Finally, bag-of-words classifiers are strong baselines for sentiment and when combined with other features can still give state-of-the-art results for sentiment tasks \cite{Mohammad2013}. Therefore, we train a Linear SVM on a bag-of-words representation of the training sentences.

\begin{table*}

\newcommand{\sep}{\cmidrule(lr){3-3}\cmidrule(lr){4-4}\cmidrule(lr){5-5}\cmidrule(lr){6-6}\cmidrule(lr){7-7}\cmidrule(lr){8-8}}

\centering\small
\begin{tabular}{llcccccc}
\toprule
		&& MPQA & OpeNER & SemEval & SST & Täckström & Thelwall\\ 
\sep
\multirow{4}{*}{\rt{Single}}
& BOW	  & 40.9 & 69.7 & 62.3 & 50.9 & 46.0 & 53.5 \\
& BiLSTM  &	48.7 & 71.5 & 58.0 & 37.5 & 45.0 & 52.0 \\
& ELMo    &	61.0 & 82.1 & 71.9 & 51.3 & 53.1 & 59.1 \\
& BERT    &	\textbf{62.3} & \textbf{84.2} & \textbf{75.1} & \textbf{53.0} & \textbf{60.2} & \textbf{63.9} \\

\bottomrule
\end{tabular}
\caption{Accuracy of models on the sentiment datasets, where a different classifier is trained for each dataset.}
\label{table:results}
\end{table*}

\subsection{Model performance}

Table \ref{table:results} shows the accuracy of the models on the six tasks. Both methods that use pretrained language model classifiers (ELMo and BERT) are the best performing models, with an average of 11.8 difference between the language model classifiers and standard models (BOW and BILSTM). The error rates range between 8.3 on OpeNER and 20.5 on SST (see Table \ref{table:errors}), indicating that there are differences in difficulty of datasets due to domain and annotation characteristics.

Additional experiments on a \textsc{Merged} setup, where the labels from OpeNER and SST are mapped to the three-class setup, and a single model is trained on the concatenation of the training sets from all datasets, indicate that no clear performance gain is achieved. We therefore prefer to avoid the problem of domain differences and keep only the original results.

\section{Challenging Dataset}
\label{challengedataset}

We create a challenging dataset by collecting the subset of test sentences that \emph{all} of the sentiment systems predicted incorrectly (statistics are shown in Table \ref{table:errors}). After removing sentences with incorrect gold labels, there are a total of 836 sentences in the dataset, with a similar number of positive, neutral, and negative labels and fewer strong labels. This is expected, as only two datasets have strong labels.

Furthermore, the main sources of examples are the SemEval task (249), Stanford Sentiment Treebank (452) and Thelwall datasets (215), while the Täckström dataset (129), MPQA (39) and OpeNER (29) contribute much less. This is a result of both dataset size and difficulty. 

\begin{table*}

\newcommand{\sep}{\cmidrule(lr){2-2}\cmidrule(lr){3-3}\cmidrule(lr){4-4}\cmidrule(lr){5-5}\cmidrule(lr){6-6}\cmidrule(lr){7-7}\cmidrule(lr){8-8}}

\centering\small
\begin{tabular}{lrrrrrr|r}
\toprule
		& MPQA & OpeNER & SemEval & SST & Täckström & Thelwall & Total\\ 

\sep
$++$ & $-$ & 8   & $-$ & 87  & $-$ & $-$ &  95  \\
$+$  & 16  & 9   & 59  & 49  & 46  & 9   &  188 \\
0    & 1   & $-$ & 45  & 75  & 31  & 48  &  200 \\
$-$  & 16  & 2   & 47  & 51  & 18  & 116 &  250 \\
$--$ & $-$ & 4   & $-$ & 99  & $-$ & $-$ &  103 \\
\sep
 Total & 33 & 23 & 151 & 361 & 95 & 173 & \textbf{836}\\
 \% of original & 14.5  & 6.6 & 6.4 & 16.3 & 12.9 & 13.6  & 11.7\\
  avg. length & 25.0 & 13.4 & 19.0 & 19.9 & 23.4 & 17.5 & 19.7\\
\bottomrule
\end{tabular}
\caption{Statistics of dataset, including the number of sentences from each dataset and for each label, the percentage of the original dataset kept in the dataset, and average length (in tokens) of sentences.}
\label{table:errors}
\end{table*}

\begin{table}[b]
\begin{tabular}{ll}
\toprule
Strong Positive & It was \underline{\textbf{spot on}}. \\
Positive & They're \underline{\textbf{on a roll}}. \\
Neutral & It's a bit \underline{\textbf{hit-or-miss}}.\\
Negative & I'm \underline{\textbf{pulling my hair out}}. \\
Strong Negative & Madonna \underline{\textbf{can't act a lick}}.\\
\bottomrule
\end{tabular}
\caption{Examples of idioms.}
\label{setphrase:examples}
\end{table}

\section{Dataset Analysis}
\label{datasetanalysis}

In order to give a clearer view of the data found in the dataset, we annotate these instances using 19 linguistic and paralinguistic labels. While most of these come from previous attempts to qualitatively analyze sentiment classifiers \cite{HuandLiu2004,Das2007,PangLee2008,Socher2013b,Barnes2018a}, others (incorrect label, no sentiment, morphology) emerged during the error annotation process. We further chose to manually annotate for the polarity of the sentence irrespective of the gold label in order to be able to locate possible annotation errors during our analysis. The annotation scheme and (manually constructed) examples of each label are shown in Table \ref{error_labels}. Note that we did not limit the number of labels that the annotator could assign to each sentence and in principle they should assign all suitable labels during annotation.

\begin{figure}[h!]
\centering
\includegraphics[scale=0.7]{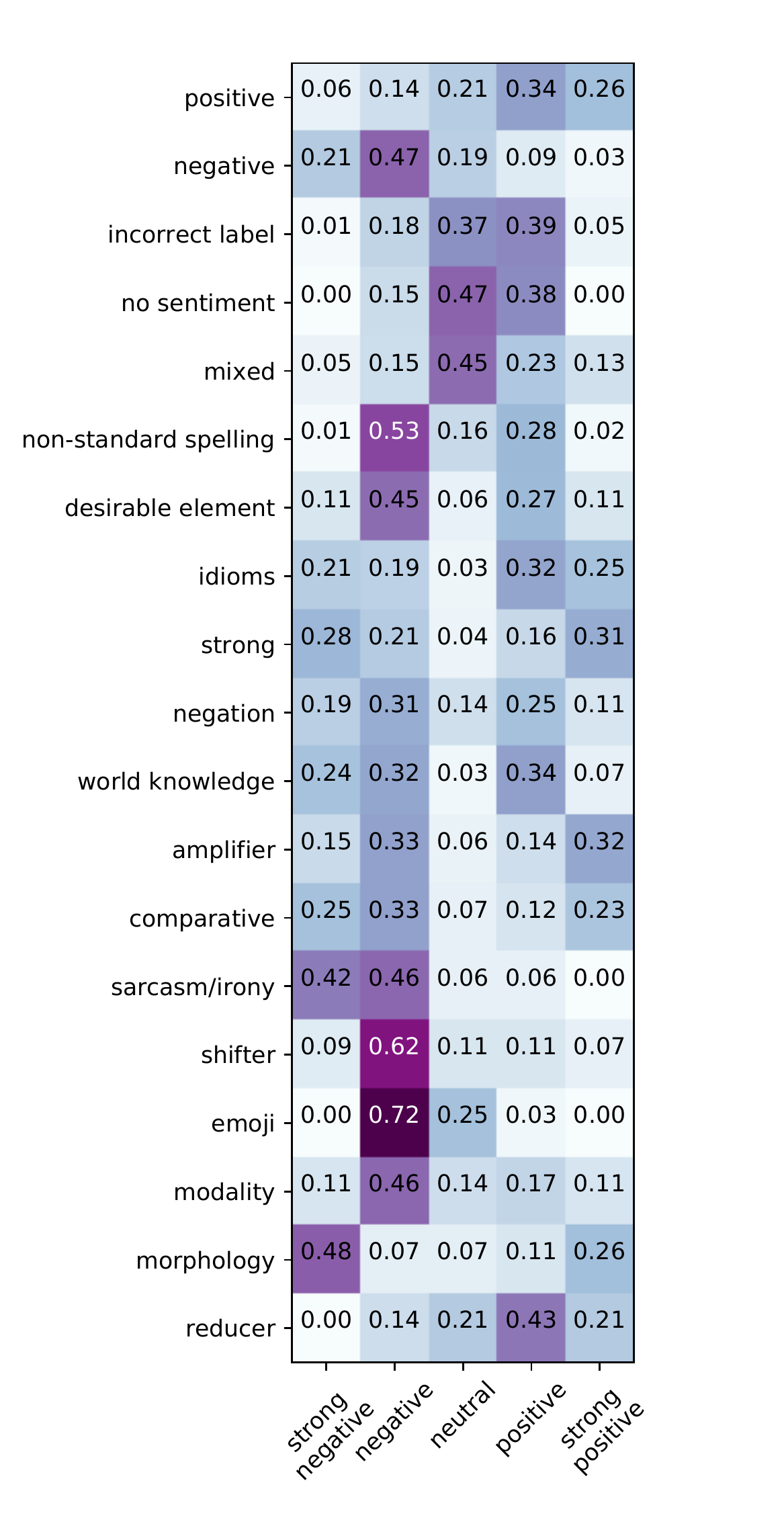}
\caption{Distribution of labels across error categories.}
\label{fig:errordist}
\end{figure}

\begin{table}[t]
\centering
\begin{tabular}{lc}
\toprule
label & \# examples \\
\cmidrule(lr){1-1}\cmidrule(lr){2-2}
incorrect label & \textbf{277} \\
no sentiment & \textbf{214} \\
mixed & \textbf{185}\\
non-standard spelling & \textbf{180}\\
desirable element &  \textbf{144}\\
idioms & 132 \\
strong & 122\\
negation &  97\\
world knowledge  & 81\\
amplifier &  79\\
comparative & 68\\
sarcasm/irony & 58\\
shifter & 50\\
emoji & 46 \\
modality &  38\\
morphology &  31\\
reducer &  13\\
\bottomrule
\end{tabular}

\caption{Number of labels for each category in annotation study. \textbf{Bold} numbers indicate the five most frequent sources of errors. The total number of labels does not sum to the number of sentences in the dataset, as each sentence can have multiple labels.}
\label{errorcategories}
\end{table}

An initial analysis of the errors shown in Table \ref{errorcategories} and Figure \ref{fig:errordist} reveals that the most common errors come from the no-sentiment (214), mixed category (185), non-standard spelling and hashtags (180), desirable elements (144), and the strong label (122).

The distribution of errors across labels (strong negative: 106, negative: 299, neutral: 303, positive: 296, strong positive: 109) compared to the gold distribution (strong negative: 294, negative: 1742, neutral: 2249, positive: 2402, strong positive: 475) shows that the strong negative is the most difficult and least common class, while positive is the easiest to classify. In the following we briefly discuss the error categories, also showing examples for each.

\begin{table*}[t]
\begin{tabular}{ll}
\toprule

positive  & ``It was good.'' \\
negative  & ``It was bad.'' \\
negation & ``It was not good.'' \\
strong & ``It was incredible.'' \\
amplifier & ``It was really good.'' \\
reducer & ``It was kind of bad.''\\
desirable element & ``It had a pool.'' \\
comparative & ``It was better than the first hotel.'' \\
shifter & ``They denied him the scholarship'' \\
modality & ``I would have loved the room if it been bigger.'' \\
world knowledge & ``It was 2 minutes from the beach.'' vs. ``It was 2 hours from the beach.'' \\
morphology & ``It was un-fricking-believable.'' \\
non-standard spelling & ``It was awesoooome.'' \\
idioms & ``It's not my cup of tea.'' \\
sarcasm/irony & ``I love it when people yell at me first thing in the morning.'' \\
emoji & ``:)'' \\
no sentiment & ``The president will hold a talk tomorrow.''\\
mixed & ``The plot was nice, but a little slow.''\\
incorrect label & Any clearly incorrect label. \\
\bottomrule
\end{tabular}
\caption{Categories and examples for error annotation guidelines.}
\label{error_labels}
\end{table*}

\paragraph{Mixed Polarity}

The largest set of errors, with 185 sentences labeled, are what we refer to as ``mixed'' polarity sentences. These are sentences where two differing polarities are expressed, either towards two separate entities, or towards the same entity. While the first can be solved by a more fine-grained approach (aspect-level or targeted sentiment), the second is more difficult and is often considered a category of its own \cite{Shamma2009,Saif2013EvaluationDF,Kenyon-Dean2018}.

An analysis of the mixed category errors reveals that while most of the examples are in the ``neutral'' category (45\%), the other 55\% are annotated as having mostly positive or negative sentiment. This is a confusing situation for both annotators and sentiment classifiers, and a direct product of performing sentence-level classification rather than aspect-level. Nearly a third of the errors contain ``but'' clauses, which could be correctly classified by splitting them.

A more problematic situation is found among nearly 20\% of the examples (34), where the annotator found the original label to be completely incorrect.\footnote{We do not include examples where only the strength of the polarity was considered different, \ie, positive vs. strong positive.}

\paragraph{Non-standard spelling}

Most errors in this category (180 total) are labeled either negative (49\%) or positive (29\%), with almost no strong positive or strong negative, which comes mainly from the fact that the noisier datasets do not contain the strong labels.

Around a third of the examples contain hashtags that clearly express the sentiment of the whole sentence, \eg, ``\#imtiredof this SNOW and COLD weather!!!''. This indicates the need to properly deal with hashtags in order to correctly classify sentiment.

\paragraph{Idioms}

Table \ref{setphrase:examples} presents some examples of sentiment-bearing idioms that are taken from the challenge data set.
In this category, errors (132 sentences labeled) are spread relatively uniformly across labels. Learning these correctly from sentence-level annotations is unlikely, especially because they are seldom found repeatedly, even in a training corpus of decent size. Therefore, incorporating idiomatic information from external data sources may be necessary to improve the classification of sentences within this category.

\paragraph{Strong Labels}

This category (122 total) is particularly difficult for sentiment classifiers for several reasons. First, strong negative sentiment is often expressed in an understated or ironic manner. For example, ``Better at putting you to sleep than a sound machine.''

For strong positive examples in the dataset, there is often difficult vocabulary and morphologically creative uses of language, \eg, ``It is a kickass , dense sci-fi action thriller hybrid that delivers and then some.'', while strong negative examples often contain sarcasm or non-standard spelling, \eg, ``All prints of this film should be sent to and buried on Pluto.''.

\paragraph{Negation}

Negation, which accounts for 97 errors, directly affects the classification of polar sentence \cite{Wiegand2010}. Therefore, we look at the differences between correctly and incorrectly classified sentences containing negation, by analyzing 100 correctly and incorrectly classified sentences containing negation.

From our analysis, there is no specific negator that is more difficult to resolve regarding its effect on sentiment classification.

We also perform an analysis of negation scope under the assumption that when a negator occurs farther from its negated element, it is more difficult for the sentiment classifier to correctly resolve the negation. Let $d$ be the distance between the negator $n$ and the relevant sentiment element $se$, such that $d = |ind(se) - ind(n)|$ where the function $ind$ calculates the index of a token in a sentence. We find that the incorrectly classified examples have an average $d$ of 2.7, while the correctly classified examples had 2.5. This seems to rule out a problem of negation scope as the underlying difference.

High-level or clausal negation occurs when the negator negates a full clause, rather than an adjective or noun phrase, \eg, ``I don't think it is a particularly interesting film''. In the dataset this phenomenon is found more prevalently in the incorrectly classified examples (8\%) versus the correctly classified examples (3\%), but does not occur often in absolute terms.

The main source of difference regarding correctly classifying examples involving negation seems to be irrelevant negation. Irrelevant negation refers to cases where a sentence contains a negation but where the sentiment-bearing expression is not within the scope of negation. In our data, there is a strong difference in the distribution of irrelevant negation in correctly and incorrectly classified examples (80\% vs. 25\%, respectively), suggesting that sentiment classifiers learn to ignore most occurrences of negation.

\paragraph{World Knowledge}

Examples from the dataset where world knowledge is necessary to correctly classify a sentence (81 sentences) include comparisons with entities commonly associated with positive or negative polarity, \eg, ``Elicits more groans from the audience than Jar Jar Binks, Scrappy Doo and Scooby Dumb, all wrapped up into one.'', analogies, \eg, ``Adam Sandler is to Gary Cooper what a gnat is to a racehorse.'', or rating scales, \eg, ``10/10 overall''. 

This category is also highly correlated with sarcasm and irony. In fact, irony is often defined as ``violating expectations'' \cite{Hao2010}, which presupposes that we possess a world knowledge containing expectations of a situation.

\paragraph{Amplified}

Amplifiers occur mainly in negative and strong positive examples, such as ``It's an awfully derivative story.'' Most of the amplified sentences found in the dataset (71/79) contain amplifiers other than ``very'', such as ``super'', ``incredibly'', or ``so''.

\paragraph{Comparative}

Comparative sentiment, with 68 errors, is known to be difficult \cite{HuandLiu2004,Liu2012}, as it is necessary to determine which entity is on which side of the inequality. Sentences like ``Will probably stay in the shadow of its two older, more accessible Qatsi siblings'' are difficult for sentiment classifiers that do not model this phenomenon explicitly.

\paragraph{Sarcasm/Irony}

Sarcasm and irony (58 errors), which are often treated separately from sentiment analysis \cite{Filatova2012,Barbieri2014}, are present mainly in negative and strong negative examples in the dataset. Correctly capturing sarcasm and irony is necessary to classify some negative and strong negative examples, \eg, ``If Melville is creatively a great whale, this film is canned tuna.''

\paragraph{Shifters}

Shifters (50 errors), such as ``abandon'', ``lessen'', or ``reject'' are less common within the dataset, but normally move positive polarity words towards a more negative sentiment. The most common shifter is the word ``miss'', used as in ``We miss the quirky amazement that used to come along for an integral part of the ride.''

\paragraph{Emoji}

While the models handle most occurrences of emojis well, they falter more on the negative examples (46 errors). More than half of the examples in the dataset present positive emoji with a negative gold label, such as ``Pricess Leia is going to be gutted! :-).''

\paragraph{Modality}

None of the state-of-the-art sentiment systems deals explicitly with modality (38 total errors). While in many of the examples modality does not express a different sentiment than the same sentence without modality, in the dataset there are examples that do, \eg, ``Still, I thought it could have been more.''

\paragraph{Morphology}

While not the most prominent label (31 errors), the examples in the dataset that contain morphological features that effect sentiment are normally strong positive or strong negative. This most often contains creative use of English morphology, \eg, ``It was fan-freakin-tastic!'' or ``It's hyper-cliched''.

\paragraph{Reducers}

Reducers (13 errors), such as ``kind of'', ``less'', or ``all that'' cooccur with both positive and negative polar words within the dataset, and tend to lead to positive or neutral sentiment, \eg, ``It was a lot less hassle.''

\section{Case Study: Training with phrase-level annotations}
\label{casestudy}

As a case study for the usage of the dataset presented here, we evaluate a model that has access to more compositional information. Besides having sentence-level annotations, the SST dataset also contains annotations for each phrase in a constituency tree, which gives a considerable amount more training data, specifically 155,019 annotated phrases vs. 8,544 annotated sentences. It has been claimed that this data allows models to learn more compositionality \cite{Socher2013b}. Therefore, we fine-tune the best performing model (BERT) on this data and test on our dataset. The BERT model trained on phrases achieves 55.1 accuracy on the SST dataset, versus 53.0 for the model trained only on sentence-level annotations. 

\begin{table}
\centering

\newcommand{\pointout}[1]{\textbf{#1}}

\begin{tabular}{llll}
\toprule
label & Sent. & Phrases & Rel. Imp.\\
\cmidrule(lr){1-1}\cmidrule(lr){2-2}\cmidrule(lr){3-3}\cmidrule(lr){4-4}
overall      &   23.0    & 31.1 & 10.5\\   
\cmidrule(lr){1-1}\cmidrule(lr){2-2}\cmidrule(lr){3-3}\cmidrule(lr){4-4}
positive     & 19.0 & \pointout{26.9} & 9.8\%\ \\  
negative     & 23.1 & \pointout{35.0} & 15.5\% \\ 
mixed        & 21.2 & \pointout{26.5} & 6.7\%  \\
no-sentiment & 37.6 & \pointout{42.6} & 8.1\%  \\ 
non-strd spelling     & 40.3 & \pointout{43.5} & 3.8\% \\ 
desirable    & 25.7 & \pointout{28.7} & 4.0\%  \\ 
idioms   & 13.7 & \pointout{23.1} & 11.0\% \\ 
strong       & 15.5 & \pointout{23.7} & 9.7\%  \\ 
negation     & 23.9 & \pointout{38.6} & \posbox{19.3\%} \\ 
world know.  & 14.9 & \pointout{21.6} & \posbox{19.6\%} \\ 
amplified    & 13.9& \pointout{31.9}  & \posbox{20.9\%} \\  
comparative  & 11.7 & \pointout{13.3} & 1.8\%  \\ 
irony        & \pointout{20.8} & 18.8 & \negbox{-2.5\%} \\ 
shifters     & \pointout{33.3} & 24.4 & \negbox{-11.8\%}\\ 
emoji        & 33.3 & \pointout{50.0} & \posbox{25.0\%} \\ 
modality  & 20 & \pointout{22.9} & 3.6\%\\ 
morphology       & \pointout{18.5}  & \pointout{18.5} & \negbox{0\%}\\ 
reduced      & 7.7  & \pointout{23.1} & \posbox{16.7\%} \\

\bottomrule
\end{tabular}
\caption{Per category accuracy and relative improvement (last column) of BERT model trained on SST sentences (8,544) and SST phrases (155,019).}
\label{table:casestudy}
\end{table}

Table \ref{table:casestudy} shows that the model trained on the SST phrases performs overall much better than the model trained on SST sentences\footnote{It is important to realize that the SST-sentence model has 0 accuracy on the subset of the dataset taken from the SST dataset, but not on the sentences taken from the other datasets.} on the dataset. Using the error annotations in the challenge data set, we find that results improve greatly on the sentences which contain the labels negation, world knowledge, amplified, emoji, and reduced, while performing worse on irony, shifters and equally on morphology. This analysis seems to indicate that phrase-level annotations help primarily with learning compositional sentiment (negation, amplified, reduced), while other phenomena, such as irony or morphology do not receive improvements. This confirms that training on the phrase-level annotations improves a sentiment model's ability to classify compositional sentiment, while also demonstrating the usefulness of our dataset for introspection.

\section{Conclusion and Future Work}
\label{conclusion}

In this paper, we tested three state-of-the-art sentiment classifiers and a baseline bag-of-words classifier on six English sentence-level sentiment datasets. We gathered the sentences that all methods misclassified in order to create a dataset. Additionally, we performed a fine-grained annotation of error types in order to provide insight into the kinds of problems sentiment classifiers have. We will release both the code and the annotated data with the hope that future research will utilize this resource
to probe sentiment classifiers for qualitative differences, rather than rely only on quantitative scores, which often obscure the plentiful challenges that still exist.

Many of the phenomena found in the dataset, \eg, negation or modality, have been discussed in depth in \cite{Liu2012}. However, the dataset that resulted from this work demonstrates that modern neural methods still fail on many examples of these phenomena. Additionally, our dataset enables a quick analysis of qualitative differences between models, probing their performance with respect to the linguistic and paralinguistic categories of errors. 

Additionally, many of the findings from this paper are likely to vary to a degree for other languages, due to typological differences, as well as differences in available training data. The annotation method proposed in this paper, however, should enable the creation of similar analyses and datasets in other languages.

We expect that this approach to creating a dataset is also easily transferable to other tasks which are affected by linguistic or paralinguistic phenomena, such as hate speech detection or sarcasm detection. It would be more useful to have some knowledge of the phenomena that could affect the task beforehand, but a careful error analysis can also lead to insights which can be translated into annotation labels.

Regarding ways of moving forward, there are already many sources of data for the linguistic phenomena we have analyzed in this work, ranging from datasets annotated for negation \cite{Morante2012,Liu:Fan:Web:18}, irony \cite{van-hee-etal-2018-semeval}, emoji \cite{barbieri-etal-2018-semeval}, as well as datasets for idioms \cite{muzny-zettlemoyer-2013-automatic} and their relationship with sentiment \cite{jochim-etal-2018-slide}. We believe that discovering ways to explicitly incorporate this available information into state-of-the-art sentiment models may provide a way to improve current approaches. Multi-task learning \cite{Caruana93multitasklearning} and transfer learning \cite{Peters2018,Devlin2018,Howard2018} have shown promise in this respect, but have not been exploited for improving sentiment classification with regards to these specific phenomena.

\section*{Acknowledgements}
This work has been carried out as part of the SANT project, funded by the Research Council of Norway (grant number 270908).

\bibliography{lit}
\bibliographystyle{acl_natbib}

\end{document}